\documentclass[11pt]{article}
\usepackage{colacl}

\title{Conditions on Consistency of\\Probabilistic Tree Adjoining
  Grammars\thanks{This research was partially supported by NSF grant
    SBR8920230 and ARO grant DAAH0404-94-G-0426. The author would like
    to thank Aravind Joshi, Jeff Reynar, Giorgio Satta, B. Srinivas,
    Fei Xia and the two anonymous reviewers for their valuable
    comments.}}

\author{Anoop Sarkar \\
Dept. of Computer and Information Science \\
University of Pennsylvania \\
200 South 33rd Street, \\
Philadelphia, PA 19104-6389 USA \\
{\tt anoop@linc.cis.upenn.edu}}

\usepackage{graphicx}
\usepackage{psfrag}

\newcommand{\comment}[1]{}

\newcommand{\myterm}{{\Sigma}}
\newcommand{\nonterm}{N}
\newcommand{\myvar}{V}

\newcommand{\ep}{\epsilon}

\newcommand{\expmtx}{{\cal M}}
\newcommand{\pmtx}{{\mathbf P}}
\newcommand{\nmtx}{{\mathbf N}}

\newcommand{\adj}[2]{#1 \mapsto #2}

\newcommand{\anodes}[1]{\mbox{${\cal A}(#1)$}}
\newcommand{\adjoinable}[1]{\mbox{{\em Adj}$(#1)$}}

\newcommand{\lab}{\mbox{{\it label\/}}}

\newtheorem{thm}{Theorem}[section]

\begin{document}
\maketitle

\begin{abstract}
  Much of the power of probabilistic methods in modelling language
  comes from their ability to compare several derivations for the same
  string in the language. An important starting point for the study of
  such cross-derivational properties is the notion of {\em
    consistency}. The probability model defined by a probabilistic
  grammar is said to be {\em consistent} if the probabilities assigned
  to all the strings in the language sum to one. From the literature
  on probabilistic context-free grammars (CFGs), we know precisely the
  conditions which ensure that consistency is true for a given CFG.
  This paper derives the conditions under which a given probabilistic
  Tree Adjoining Grammar (TAG) can be shown to be consistent. It gives
  a simple algorithm for checking consistency and gives the formal
  justification for its correctness. The conditions derived here can
  be used to ensure that probability models that use TAGs can be
  checked for {\em deficiency} (i.e.\ whether any probability mass is
  assigned to strings that cannot be generated).
\end{abstract}

\section{Introduction}

Much of the power of probabilistic methods in modelling language comes
from their ability to compare several derivations for the same string
in the language. This cross-derivational power arises naturally from
comparison of various derivational paths, each of which is a product
of the probabilities associated with each step in each derivation.  A
common approach used to assign structure to language is to use a
probabilistic grammar where each elementary rule or production is
associated with a probability. Using such a grammar, a probability for
each string in the language is computed.  Assuming that the
probability of each derivation of a sentence is well-defined, the
probability of each string in the language is simply the sum of the
probabilities of all derivations of the string. In general, for a
probabilistic grammar $G$ the language of $G$ is denoted by $L(G)$.
Then if a string $v$ is in the language $L(G)$ the probabilistic
grammar assigns $v$ some non-zero probability.

There are several cross-derivational properties that can be studied for
a given probabilistic grammar formalism. An important starting point
for such studies is the notion of {\em consistency}. The probability
model defined by a probabilistic grammar is said to be {\em
  consistent} if the probabilities assigned to all the strings in the
language sum to $1$. That is, if $\Pr$ defined by a probabilistic
grammar, assigns a probability to each string $v \in \myterm^\ast$,
where $\Pr(v) = 0$ if $v \not\in L(G)$, then
\begin{eqnarray}
  \label{eqn:consistency}
  \sum_{v \in L(G)} \Pr(v) = 1
\end{eqnarray}

From the literature on probabilistic context-free grammars (CFGs) we
know precisely the conditions which ensure that
(\ref{eqn:consistency}) is true for a given CFG. This paper derives
the conditions under which a given probabilistic TAG can be shown to
be consistent. 

TAGs are important in the modelling of natural language since they can
be easily lexicalized; moreover the trees associated with words can be
used to encode argument and adjunct relations in various syntactic
environments. This paper assumes some familiarity with the TAG
formalism. \cite{Joshi.MOLbook} and
\cite{Joshi.Schabes.TreeAutomatabook} are good introductions to the
formalism and its linguistic relevance. TAGs have been shown to have
relations with both phrase-structure grammars and dependency grammars
\cite{Rambow.Joshi.MTTbook} and can handle (non-projective) long
distance dependencies.

Consistency of probabilistic TAGs has practical significance for the
following reasons:
\begin{itemize}
\item The conditions derived here can be used to ensure that
  probability models that use TAGs can be checked for {\em
    deficiency}.
\item Existing EM based estimation algorithms for probabilistic TAGs
  assume that the property of consistency holds
  \cite{Schabes.Coling92}. EM based algorithms begin with an initial
  (usually random) value for each parameter. If the initial assignment
  causes the grammar to be inconsistent, then iterative re-estimation
  might converge to an inconsistent grammar\footnote{ Note that for
    CFGs it has been shown in~\cite{Chaudhari.etal,Sanchez.Benedi}
    that inside-outside reestimation can be used to avoid
    inconsistency. We will show later in the paper that the method
    used to show consistency in this paper precludes a straightforward
    extension of that result for TAGs. }.
\item Techniques used in this paper can be used to determine
  consistency for other probability models based on TAGs
  \cite{Carroll.IWPT}.
\end{itemize}

\section{Notation}
\label{sec:notation}

In this section we establish some notational conventions and
definitions that we use in this paper. Those familiar with the TAG
formalism only need to give a cursory glance through this section.

A probabilistic TAG is represented by $(\nonterm, \myterm, {\cal I},
{\cal A}, S, \phi)$ where $\nonterm, \myterm$ are, respectively,
non-terminal and terminal symbols. ${\cal I} \cup {\cal A}$ is a set
of trees termed as {\em elementary trees}. We take $\myvar$ to be the
set of all nodes in all the elementary trees. For each leaf $A \in
\myvar$, $\lab(A)$ is an element from $\myterm \cup \{ \ep \}$, and
for each other node $A$, $\lab(A)$ is an element from $\nonterm$. $S$
is an element from $\nonterm$ which is a distinguished start symbol.
The root node $A$ of every initial tree which can start a derivation
must have $\lab(A) = S$.
%

${\cal I}$ are termed {\em initial trees} and ${\cal A}$ are {\em
  auxiliary trees} which can rewrite a tree node $A \in \myvar$. This
rewrite step is called {\bf adjunction}.  $\phi$ is a function which
assigns each adjunction with a probability and denotes the set of
parameters in the model. In practice, TAGs also allow a leaf nodes $A$
such that $\lab(A)$ is an element from $\nonterm$. Such nodes $A$ are
rewritten with initial trees from ${\cal I}$ using the rewrite step
called {\bf substitution}. Except in one special case, we will not
need to treat substitution as being distinct from adjunction.

For $t \in {\cal I} \cup {\cal A}$, $\anodes{t}$ are the nodes in tree
$t$ that can be modified by adjunction. For $\lab(A) \in \nonterm$ we
denote $\adjoinable{\lab(A)}$ as the set of trees that can adjoin at
node $A \in \myvar$.  The adjunction of $t$ into $N \in \myvar$ is
denoted by $\adj{N}{t}$. No adjunction at $N \in \myvar$ is denoted by
$\adj{N}{nil}$. We assume the following properties hold for every
probabilistic TAG $G$ that we consider:
\begin{enumerate}
\item $G$ is {\em lexicalized}. There is at least one leaf node $a$
  that lexicalizes each elementary tree, i.e.\ $a \in \myterm$.
\item $G$ is {\em proper}. For each $N \in V$,%
  \[ \phi(\adj{N}{nil}) + \sum_t \phi(\adj{N}{t}) = 1 \]
\item Adjunction is prohibited on the foot node of every auxiliary
  tree. This condition is imposed to avoid unnecessary ambiguity and
  can be easily relaxed.
\item There is a distinguished non-lexicalized initial tree $\tau$
  such that each initial tree rooted by a node $A$ with $\lab(A) = S$
  substitutes into $\tau$ to complete the derivation.  This ensures
  that probabilities assigned to the input string at the start of the
  derivation are well-formed.
\end{enumerate}

We use symbols $S, A, B, \ldots$ to range over $\myvar$, symbols $a,
b, c, \ldots$ to range over $\myterm$. We use $t_1, t_2, \ldots$ to
range over $I \cup A$ and $\ep$ to denote the empty string. We use
$X_i$ to range over all $i$ nodes in the grammar.

\section{Applying probability measures to Tree Adjoining Languages}
\label{sec:inconsist}

To gain some intuition about probability assignments to languages, let
us take for example, a language well known to be a tree adjoining
language:
\[ L(G) = \{ a^n b^n c^n d^n | n \geq 1 \} \]

It seems that we should be able to use a function $\psi$ to assign any
probability distribution to the strings in $L(G)$ and then expect that
we can assign appropriate probabilites to the adjunctions in $G$ such
that the language generated by $G$ has the same distribution as that
given by $\psi$. However a function $\psi$ that grows smaller by
repeated multiplication as the inverse of an exponential function
cannot be matched by any TAG because of the {\em constant growth}
property of TAGs (see \cite{VijayShanker.thesis}, p. 104). An example
of such a function $\psi$ is a simple Poisson distribution
(\ref{eqn:poisson}), which in fact was also used as the counterexample
in \cite{Booth.and.Thompson} for CFGs, since CFGs also have the
constant growth property.
\begin{eqnarray}
  \label{eqn:poisson}
  \psi(a^n b^n c^n d^n) = \frac{1}{e \cdot n!}
\end{eqnarray}
This shows that probabilistic TAGs, like CFGs, are constrained in the
probabilistic languages that they can recognize or learn. As shown
above, a probabilistic language can fail to have a generating
probabilistic TAG. 

The reverse is also true: some probabilistic TAGs, like some CFGs,
fail to have a corresponding probabilistic language, i.e.\ they are not
consistent. There are two reasons why a probabilistic TAG could be
inconsistent: ``dirty'' grammars, and destructive or incorrect
probability assignments.

{\bf ``Dirty'' grammars}. Usually, when applied to language, TAGs are
lexicalized and so probabilities assigned to trees are used only when
the words anchoring the trees are used in a derivation.  However, if
the TAG allows non-lexicalized trees, or more precisely, auxiliary
trees with no yield, then looping adjunctions which never generate a
string are possible.  However, this can be detected and corrected by a
simple search over the grammar. Even in lexicalized grammars, there
could be some auxiliary trees that are assigned some probability mass
but which can never adjoin into another tree. Such auxiliary trees are
termed {\em unreachable} and techniques similar to the ones used in
detecting unreachable productions in CFGs can be used here to detect
and eliminate such trees.

{\bf Destructive probability assignments.} This problem is a more
serious one, and is the main subject of this paper. Consider the
probabilistic TAG shown in (\ref{ex:inconsist.tag})\footnote{ The
  subscripts are used as a simple notation to uniquely refer to the
  nodes in each elementary tree. They are not part of the node label
  for purposes of adjunction. }.
\begin{eqnarray}
  \label{ex:inconsist.tag}
  \psfrag{t1}{$t_1$}
  \psfrag{t2}{$t_2$}
  \psfrag{ep}{$\ep$}
  \psfrag{S1}{$S_1$}
  \psfrag{S2}{$S_2$}
  \psfrag{S3}{$S_3$}
  \psfrag{S*}{$S\ast$}
  \psfrag{phi(S1->t2)=1.0}{$\phi(\adj{S_1}{t_2})=1.0$}
  \psfrag{phi(S2->t2)=0.99}{$\phi(\adj{S_2}{t_2})=0.99$}
  \psfrag{phi(S2->nil)=0.01}{$\phi(\adj{S_2}{nil})=0.01$}
  \psfrag{phi(S3->t2)=0.98}{$\phi(\adj{S_3}{t_2})=0.98$}
  \psfrag{phi(S3->nil)=0.02}{$\phi(\adj{S_3}{nil})=0.02$}
  \psfrag{a}{$a$}
  \includegraphics[height=1.5in]{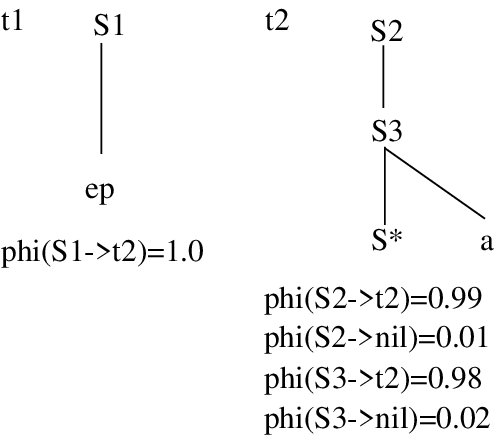}
\end{eqnarray}
Consider a derivation in this TAG as a generative process. It proceeds
as follows: node $S_1$ in $t_1$ is rewritten as $t_2$ with probability
$1.0$. Node $S_2$ in $t_2$ is $99$ times more likely than not to be
rewritten as $t_2$ itself, and similarly node $S_3$ is $49$ times more
likely than not to be rewritten as $t_2$. This however, creates two
more instances of $S_2$ and $S_3$ with same probabilities. This
continues, creating multiple instances of $t_2$ at each level of the
derivation process with each instance of $t_2$ creating two more
instances of itself. The grammar itself is not malicious; the
probability assignments are to blame. It is important to note that
inconsistency is a problem even though for any given string there are
only a finite number of derivations, all halting. Consider the
probability mass function ({\em pmf}) over the set of all derivations
for this grammar.  An inconsistent grammar would have a {\em pmf}
which assigns a large portion of probability mass to derivations that
are non-terminating.  This means there is a finite probability the
generative process can enter a generation sequence which has a finite
probability of non-termination.

\section{Conditions for Consistency}
\label{sec:conditions}

A probabilistic TAG $G$ is {\em consistent} if and only if:
\begin{eqnarray}
  \label{eqn:consistency.again}
  \sum_{v \in L(G)} \Pr(v) = 1
\end{eqnarray}
where $\Pr(v)$ is the probability assigned to a string in the
language. If a grammar $G$ does not satisfy this condition, $G$ is
said to be inconsistent.

To explain the conditions under which a probabilistic TAG is
consistent we will use the TAG in (\ref{ex:example.tag}) as an
example.
\begin{eqnarray}
  \label{ex:example.tag}
  \psfrag{t1:}{$t_1$}
  \psfrag{t2:}{$t_2$}
  \psfrag{t3:}{$t_3$}
  \psfrag{phi(A1->t2)=0.8}{$\phi(\adj{A_1}{t_2})=0.8$}
  \psfrag{phi(A1->nil)=0.2}{$\phi(\adj{A_1}{nil})=0.2$}
  \psfrag{phi(A2->t2)=0.2}{$\phi(\adj{A_2}{t_2})=0.2$}
  \psfrag{phi(A2->nil)=0.8}{$\phi(\adj{A_2}{nil})=0.8$}
  \psfrag{phi(B1->t3)=0.2}{$\phi(\adj{B_1}{t_3})=0.2$}
  \psfrag{phi(B1->nil)=0.8}{$\phi(\adj{B_1}{nil})=0.8$}
  \psfrag{phi(A3->t2)=0.4}{$\phi(\adj{A_3}{t_2})=0.4$}
  \psfrag{phi(A3->nil)=0.6}{$\phi(\adj{A_3}{nil})=0.6$}
  \psfrag{phi(B2->t3)=0.1}{$\phi(\adj{B_2}{t_3})=0.1$}
  \psfrag{phi(B2->nil)=0.9}{$\phi(\adj{B_2}{nil})=0.9$}
  \includegraphics[height=2.5in]{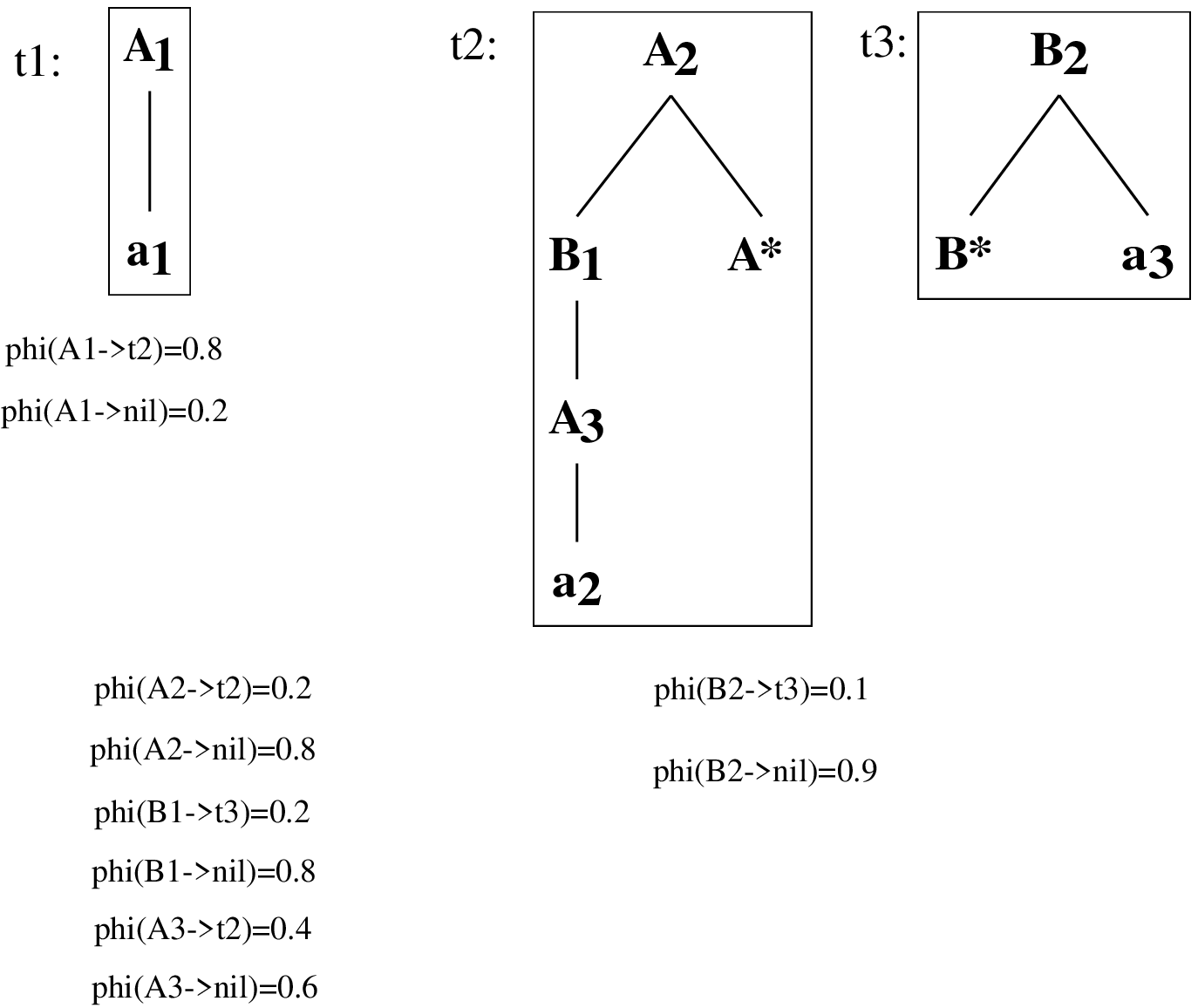}
\end{eqnarray}
From this grammar, we compute a square matrix $\expmtx$ which of size
$|V|$, where $V$ is the set of nodes in the grammar that can be
rewritten by adjunction. Each $\expmtx_{ij}$ contains the expected
value of obtaining node $X_j$ when node $X_i$ is rewritten by
adjunction at each level of a TAG derivation. We call $\expmtx$ the
stochastic {\em expectation matrix} associated with a probabilistic
TAG.

To get $\expmtx$ for a grammar we first write a matrix $\pmtx$ which
has $|V|$ rows and $|I \cup A|$ columns. An element $\pmtx_{ij}$
corresponds to the probability of adjoining tree $t_j$ at node $X_i$,
i.e.\ $\phi(\adj{X_i}{t_j})$\footnote{ Note that $\pmtx$ is not a row
  stochastic matrix. This is an important difference in the
  construction of $\expmtx$ for TAGs when compared to CFGs. We will
  return to this point in \S\ref{sec:galton.watson}. }.

\[ \pmtx = 
\begin{array}{c}
A_1 \\ A_2 \\ B_1 \\ A_3 \\ B_2
\end{array}
\stackrel{
\begin{array}{ccc}
  t_1 & t_2 & t_3
\end{array}}
{\left[
\begin{array}{rrrrr}
0 & 0.8 & 0  \\
0 & 0.2 & 0  \\
0 & 0 & 0.2  \\
0 & 0.4 & 0  \\
0 & 0 & 0.1  \\
\end{array}
\right]}
\]

We then write a matrix $\nmtx$ which has $|I \cup A|$ rows and $|V|$
columns. An element $\nmtx_{ij}$ is $1.0$ if node $X_j$ is a node in
tree $t_i$.

\[ \nmtx =
\begin{array}{c}
t_1 \\ t_2 \\ t_3
\end{array}
\stackrel{
  \begin{array}{ccccc}
    A_1 & A_2 & B_1 & A_3 & B_2
  \end{array}}
{\left[
    \begin{array}{rrrrr}
      1.0 & 0 & 0 & 0 & 0 \\
      0 & 1.0 & 1.0 & 1.0 & 0 \\
      0 & 0 & 0 & 0 & 1.0
    \end{array}
  \right]}
\]

Then the stochastic expectation matrix $\expmtx$ is simply the product
of these two matrices. 

\[ \expmtx = \pmtx \cdot \nmtx =
\begin{array}{c}
  A_1 \\ A_2 \\ B_1 \\ A_3 \\ B_2
\end{array}
\stackrel{
  \begin{array}{ccccc}
    A_1 & A_2 & B_1 & A_3 & B_2
  \end{array}}
{\left[
    \begin{array}{rrrrr}
      0 & 0.8 & 0.8 & 0.8 & 0 \\
      0 & 0.2 & 0.2 & 0.2 & 0 \\
      0 & 0 & 0 & 0 & 0.2 \\
      0 & 0.4 & 0.4 & 0.4 & 0 \\
      0 & 0 & 0 & 0 & 0.1
    \end{array}
  \right]}
\]

By inspecting the values of $\expmtx$ in terms of the grammar
probabilities indicates that $\expmtx_{ij}$ contains the values we
wanted, i.e.\ expectation of obtaining node $A_j$ when node $A_i$ is
rewritten by adjunction at each level of the TAG derivation process. 

By construction we have ensured that the following theorem from
\cite{Booth.and.Thompson} applies to probabilistic TAGs. A formal
justification for this claim is given in the next section by showing a
reduction of the TAG derivation process to a multitype Galton-Watson
branching process \cite{Harris.book}.

\begin{thm}
  A probabilistic grammar is consistent if the {\em spectral radius}
  $\rho(\expmtx) < 1$, where $\expmtx$ is the stochastic expectation
  matrix computed from the grammar. \cite{Booth.and.Thompson,Soule}
  \label{thm:booth.and.thompson}
\end{thm}

This theorem provides a way to determine whether a grammar is
consistent. All we need to do is compute the spectral radius of the
square matrix $\expmtx$ which is equal to the modulus of the largest
eigenvalue of $\expmtx$. If this value is less than one then the
grammar is consistent%
\footnote{ The grammar may be consistent when the spectral radius is
  exactly one, but this case involves many special considerations and
  is not considered in this paper. In practice, these complicated
  tests are probably not worth the effort. See \cite{Harris.book} for
  details on how this special case can be solved.}. Computing
consistency can bypass the computation of the eigenvalues for
$\expmtx$ by using the following theorem by Ger\v{s}gorin (see
\cite{Horn.and.Johnson,Wetherell}).

\begin{thm}
  For any square matrix $\expmtx$, $\rho(\expmtx) < 1$ if and only if
  there is an $n \geq 1$ such that the sum of the absolute values of
  the elements of each row of $\expmtx^n$ is less than one. Moreover,
  any $n' > n$ also has this property. (Ger\v{s}gorin, see
  \cite{Horn.and.Johnson,Wetherell})
  \label{thm:gersgorin}
\end{thm}

This makes for a very simple algorithm to check consistency of a
grammar. We sum the values of the elements of each row of the
stochastic expectation matrix $\expmtx$ computed from the grammar. If
{\em any} of the row sums are greater than one then we compute
$\expmtx^2$, repeat the test and compute $\expmtx^{2^2}$ if the test
fails, and so on until the test succeeds\footnote{ We compute
  $\expmtx^{2^2}$ and subsequently only successive powers of $2$
  because Theorem~\ref{thm:gersgorin} holds for any $n' > n$. This
  permits us to use a single matrix at each step in the algorithm. }.
The algorithm does not halt if $\rho(\expmtx) \geq 1$. In practice,
such an algorithm works better in the average case since computation
of eigenvalues is more expensive for very large matrices. An upper
bound can be set on the number of iterations in this algorithm. Once
the bound is passed, the exact eigenvalues can be computed.

For the grammar in (\ref{ex:example.tag}) we computed the following
stochastic expectation matrix:

\[ \expmtx = 
\left[
    \begin{array}{rrrrr}
      0 & 0.8 & 0.8 & 0.8 & 0 \\
      0 & 0.2 & 0.2 & 0.2 & 0 \\
      0 & 0 & 0 & 0 & 0.2 \\
      0 & 0.4 & 0.4 & 0.4 & 0 \\
      0 & 0 & 0 & 0 & 0.1
    \end{array}
  \right]
\]

The first row sum is $2.4$. Since the sum of each row must be less
than one, we compute the power matrix $\expmtx^2$. However, the sum of
one of the rows is still greater than $1$. Continuing we compute
$\expmtx^{2^2}$.

\[ \expmtx^{2^2} = 
\left[
    \begin{array}{rrrrr}
         0  &  0.1728  &   0.1728  &   0.1728   &  0.0688 \\
         0  &  0.0432  &   0.0432  &   0.0432   &  0.0172 \\
         0  &       0  &        0  &        0   &  0.0002 \\
         0  &  0.0864  &   0.0864  &   0.0864   &  0.0344 \\
         0  &     0   &       0   &       0    & 0.0001 
    \end{array}
  \right]
\]

This time all the row sums are less than one, hence $\rho(\expmtx) <
1$. So we can say that the grammar defined in (\ref{ex:example.tag})
is consistent. We can confirm this by computing the eigenvalues for
$\expmtx$ which are $0, 0, 0.6, 0$ and $0.1$, all less than $1$.

Now consider the grammar (\ref{ex:inconsist.tag}) we had considered in
Section~\ref{sec:inconsist}. The value of $\expmtx$ for that grammar
is computed to be:

\[ \expmtx_{(\ref{ex:inconsist.tag})} = 
\begin{array}{c}
  S_1 \\ S_2 \\ S_3
\end{array}
\stackrel{
  \begin{array}{ccc}
    S_1 & S_2 & S_3
  \end{array}}
{\left[
    \begin{array}{rrr}
         0  &   1.0  &   1.0 \\
         0  &   0.99  &   0.99 \\
         0  &   0.98  &   0.98
    \end{array}
  \right]}
\]

The eigenvalues for the expectation matrix $\expmtx$ computed for the
grammar (\ref{ex:inconsist.tag}) are $0$, $1.97$ and $0$. The largest
eigenvalue is greater than $1$ and this confirms
(\ref{ex:inconsist.tag}) to be an inconsistent grammar.

\section{TAG Derivations and Branching Processes}
\label{sec:galton.watson}

To show that Theorem~\ref{thm:booth.and.thompson} in
Section~\ref{sec:conditions} holds for any probabilistic TAG, it is
sufficient to show that the derivation process in TAGs is a
Galton-Watson branching process.

A Galton-Watson branching process \cite{Harris.book} is simply a model
of processes that have objects that can produce additional objects of
the same kind, i.e.\ recursive processes, with certain properties.
There is an initial set of objects in the $0$-th generation which
produces with some probability a first generation which in turn with
some probability generates a second, and so on. We will denote by
vectors $Z_0, Z_1, Z_2, \ldots$ the $0$-th, first, second, $\ldots$
generations. There are two assumptions made about $Z_0, Z_1, Z_2,
\ldots$:

\begin{enumerate}
\item The size of the $n$-th generation does not influence the
  probability with which any of the objects in the $(n+1)$-th
  generation is produced. In other words, $Z_0, Z_1, Z_2, \ldots$ form
  a Markov chain.
\item The number of objects born to a parent object does not depend on
  how many other objects are present at the same level.
\end{enumerate}

We can associate a generating function for each level $Z_i$. The value
for the vector $Z_n$ is the value assigned by the $n$-th iterate of
this generating function. The expectation matrix $\expmtx$ is defined
using this generating function.

The theorem attributed to Galton and Watson specifies the conditions
for the probability of extinction of a family starting from its $0$-th
generation, assuming the branching process represents a family tree
(i.e, respecting the conditions outlined above). The theorem states
that $\rho(\expmtx) \leq 1$ when the probability of extinction is
$1.0$.
\begin{eqnarray}
  \label{fig:derivation.tree}
  \psfrag{level 0}{\mbox{\small level $0$}}
  \psfrag{level 1}{\mbox{\small level $1$}}
  \psfrag{level 2}{\mbox{\small level $2$}}
  \psfrag{level 3}{\mbox{\small level $3$}}
  \psfrag{level 4}{\mbox{\small level $4$}}
  \includegraphics[height=1.8in]{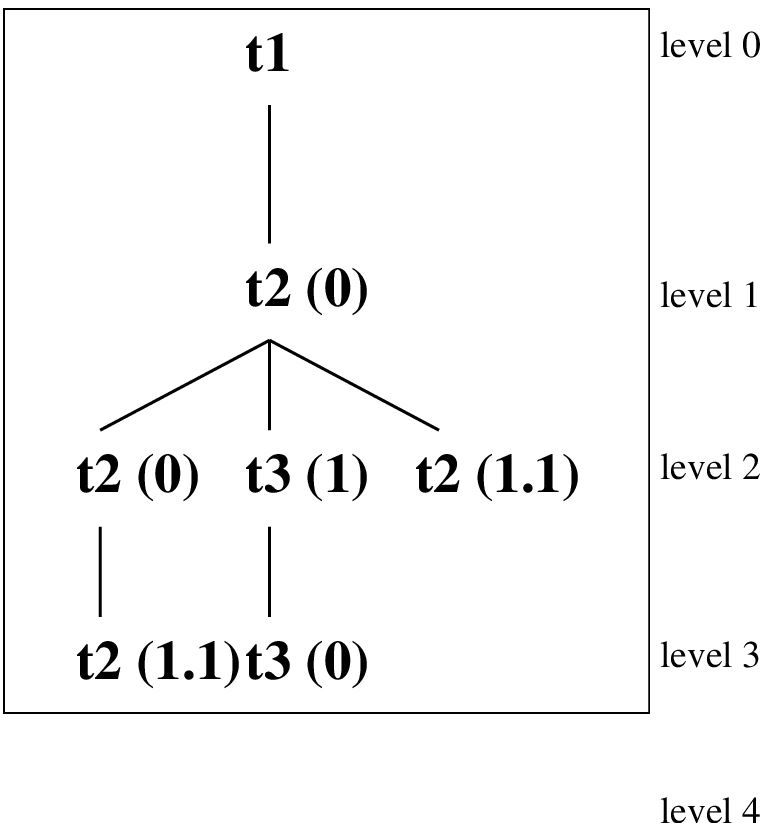} \\
  \label{fig:derived.tree}
  \includegraphics[height=3in]{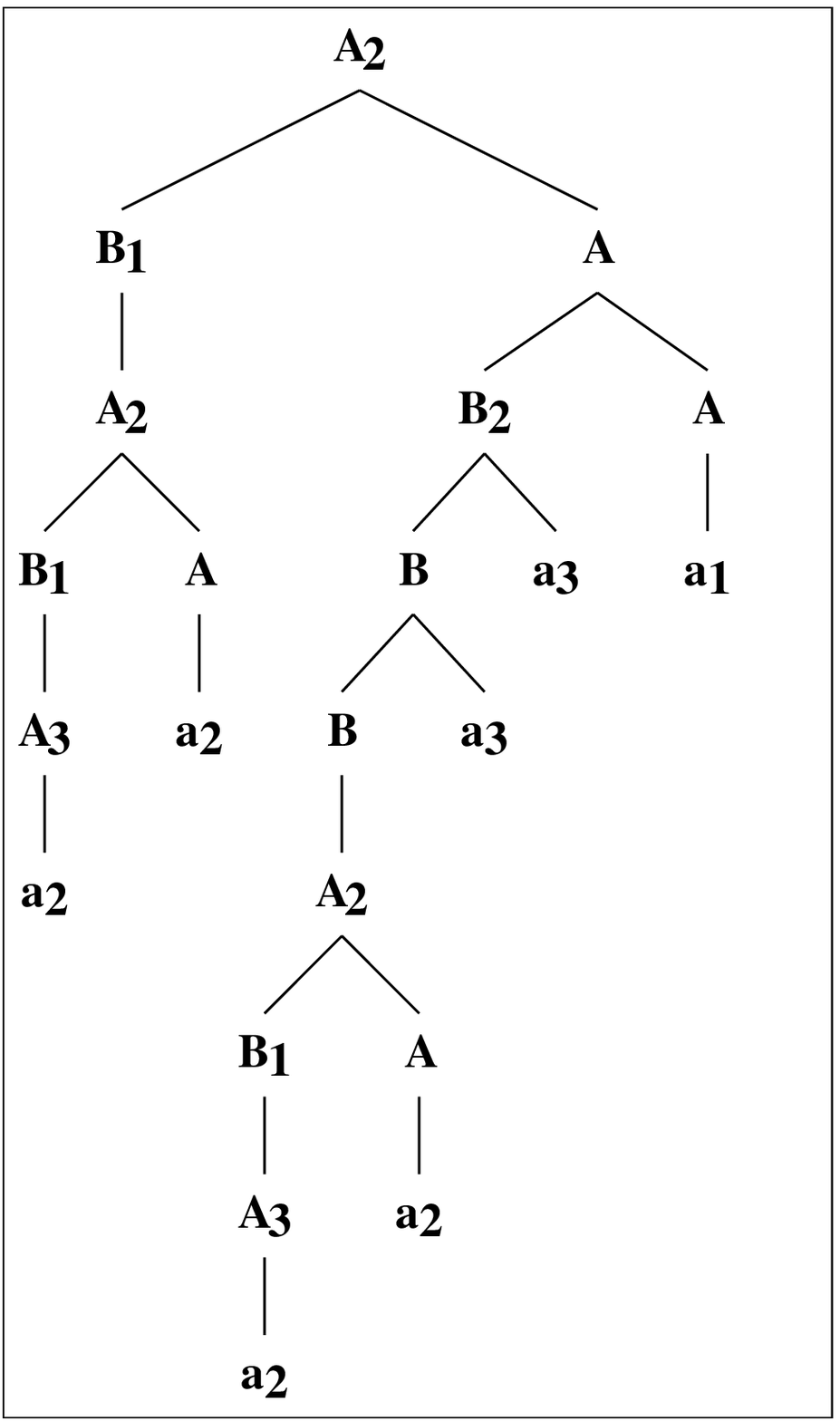}
\end{eqnarray}
The assumptions made about the generating process intuitively holds
for probabilistic TAGs. (\ref{fig:derivation.tree}), for example,
depicts a derivation of the string $a_2 a_2 a_2 a_2 a_3 a_3 a_1$ by a
sequence of adjunctions in the grammar given in
(\ref{ex:example.tag})\footnote{ The numbers in parentheses next to
  the tree names are node addresses where each tree has adjoined into
  its parent. Recall the definition of node addresses in
  Section~\ref{sec:notation}. }. The parse tree derived from such a
sequence is shown in Fig.~\ref{fig:derived.tree}. In the derivation
tree (\ref{fig:derivation.tree}), nodes in the trees at each level $i$
are rewritten by adjunction to produce a level $i+1$.  There is a
final level $4$ in (\ref{fig:derivation.tree}) since we also consider
the probability that a node is not rewritten further, i.e.\
$\Pr(\adj{A}{nil})$ for each node $A$.

We give a precise statement of a TAG derivation process by defining a
generating function for the levels in a derivation tree. Each level
$i$ in the TAG derivation tree then corresponds to $Z_i$ in the Markov
chain of branching processes. This is sufficient to justify the use of
Theorem~\ref{thm:booth.and.thompson} in Section~\ref{sec:conditions}.
The conditions on the probability of extinction then relates to the
probability that TAG derivations for a probabilistic TAG will not
recurse infinitely.  Hence the probability of extinction is the same
as the probability that a probabilistic TAG is consistent.

For each $X_j \in V$, where $V$ is the set of nodes in the grammar
where adjunction can occur, we define the $k$-argument {\em adjunction
  generating function\/} over variables $s_1, \ldots, s_k$
corresponding to the $k$ nodes in $V$.
\begin{eqnarray*}
\lefteqn{g_j(s_1, \ldots, s_k) = }\\
&& \sum_{t \in \adjoinable{X_j} \cup \{ nil \} }
\phi(\adj{X_j}{t}) \cdot s_1^{r_1(t)} \cdots s_k^{r_k(t)}
\end{eqnarray*}
where, $r_j (t) = 1$ iff node $X_j$ is in tree $t$, $r_j (t) = 0$
otherwise.

For example, for the grammar in (\ref{ex:example.tag}) we get the
following adjunction generating functions taking the variable $s_1,
s_2, s_3, s_4, s_5$ to represent the nodes $A_1, A_2, B_1, A_3, B_2$
respectively. 
\begin{eqnarray*}
\lefteqn{g_1(s_1, \ldots, s_5) = }\\
&&\phi(\adj{A_1}{t_2}) \cdot s_2 \cdot s_3 \cdot s_4 + \phi(\adj{A_1}{nil})\\
\lefteqn{g_2(s_1, \ldots, s_5) = }\\
&&\phi(\adj{A_2}{t_2}) \cdot s_2 \cdot s_3 \cdot s_4 + \phi(\adj{A_2}{nil})\\
\lefteqn{g_3(s_1, \ldots, s_5) = }\\
&&\phi(\adj{B_1}{t_3}) \cdot s_5 + \phi(\adj{B_1}{nil})\\
\lefteqn{g_4(s_1, \ldots, s_5) = }\\
&&\phi(\adj{A_3}{t_2}) \cdot  s_2 \cdot s_3 \cdot s_4 + \phi(\adj{A_3}{nil})\\
\lefteqn{g_5(s_1, \ldots, s_5) = }\\
&&\phi(\adj{B_2}{t_3}) \cdot  s_5 + \phi(\adj{B_2}{nil})
\end{eqnarray*}
The $n$-th level generating function $G_n(s_1, \ldots, s_k)$ is
defined recursively as follows.
\begin{eqnarray*}
  G_0(s_1, \ldots, s_k) & = & s_1 \\
  G_1(s_1, \ldots, s_k) & = & g_1(s_1, \ldots, s_k)\\
  G_n(s_1, \ldots, s_k) & = & G_{n-1} [ g_1(s_1, \ldots, s_k),
  \ldots, \\
  && g_k(s_1, \ldots, s_k) ]
\end{eqnarray*}
For the grammar in (\ref{ex:example.tag}) we get the following level
generating functions.
\begin{eqnarray*}
  \lefteqn{G_0(s_1, \ldots, s_5) = s_1}\\
  \lefteqn{G_1(s_1, \ldots, s_5) = g_1(s_1, \ldots, s_5)}\\
  & = & \phi(\adj{A_1}{t_2}) \cdot s_2 \cdot s_3 \cdot s_4 +
  \phi(\adj{A_1}{nil})\\ 
  & = & 0.8 \cdot s_2 \cdot s_3 \cdot s_4 + 0.2\\
  \lefteqn{G_2(s_1, \ldots, s_5) = }\\
  && \phi(\adj{A_2}{t_2}) [g_2(s_1, \ldots, s_5)] [g_3(s_1, \ldots,
  s_5)] \\ && \ \ \ [g_4(s_1, \ldots, s_5)] + \phi(\adj{A_2}{nil})\\
  & = & 0.08 s_2^2 s_3^2 s_4^2 s_5 + 0.03 s_2^2 s_3^2 s_4^2 + 0.04 s_2
  s_3 s_4 s_5 + \\ && \ \ \ 0.18 s_2 s_3 s_4 + 0.04 s_5 + 0.196 \\
\lefteqn{\ldots}
\end{eqnarray*}
Examining this example, we can express $G_i(s_1, \ldots, s_k)$ as a
sum $D_i(s_1, \ldots, s_k) + C_i$, where $C_i$ is a constant and
$D_i(\cdot)$ is a polynomial with no constant terms. A probabilistic
TAG will be consistent if these recursive equations terminate, i.e.\ iff
\[ lim_{i \rightarrow \infty} D_i(s_1, \ldots, s_k) \rightarrow 0 \]
We can rewrite the level generation functions in terms of the
stochastic expectation matrix $\expmtx$, where each element $m_{i,j}$
of $\expmtx$ is computed as follows (cf.~\cite{Booth.and.Thompson}).
\begin{eqnarray}
  \label{eqn:partial.dv}
  m_{i,j} = \left. \frac{\partial g_i (s_1, \ldots, s_k)}{\partial
      s_j} \right|_{s_1, \ldots, s_k = 1}
\end{eqnarray}
The limit condition above translates to the condition that the
spectral radius of $\expmtx$ must be less than $1$ for the grammar to
be consistent.

This shows that Theorem~\ref{thm:booth.and.thompson} used in
Section~\ref{sec:conditions} to give an algorithm to detect
inconsistency in a probabilistic holds for any given TAG, hence
demonstrating the correctness of the algorithm.

Note that the formulation of the adjunction generating function means
that the values for $\phi(\adj{X}{nil})$ for all $X \in V$ do not
appear in the expectation matrix. This is a crucial difference between
the test for consistency in TAGs as compared to CFGs. For CFGs, the
expectation matrix for a grammar $G$ can be interpreted as the
contribution of each non-terminal to the derivations for a sample set
of strings drawn from $L(G)$. Using this it was shown
in~\cite{Chaudhari.etal} and~\cite{Sanchez.Benedi} that a single step
of the inside-outside algorithm implies consistency for a
probabilistic CFG. However, in the TAG case, the inclusion of values
for $\phi(\adj{X}{nil})$ (which is essential if we are to interpret
the expectation matrix in terms of derivations over a sample set of
strings) means that we cannot use the method used
in~(\ref{eqn:partial.dv}) to compute the expectation matrix and
furthermore the limit condition will not be convergent. 

\section{Conclusion}

We have shown in this paper the conditions under which a given
probabilistic TAG can be shown to be consistent. We gave a simple
algorithm for checking consistency and gave the formal justification
for its correctness. The result is practically significant for its
applications in checking for {\em deficiency} in probabilistic TAGs.

\nocite{Chaudhari.etal}
\nocite{Sanchez.Benedi}
\bibliographystyle{acl}
{\footnotesize \bibliography{acl98final}}
\end{document}